\documentclass[a4paper]{article}

\usepackage{INTERSPEECH_v2}

\graphicspath{{fig/}}

\usepackage{booktabs}
\usepackage{tabularx}
\usepackage{multirow}
\newcommand{\mytable}{
    \centering
    \renewcommand{\arraystretch}{1.2}
    }
\newcolumntype{C}{>{\centering\arraybackslash}X}
\newcolumntype{L}{>{\raggedright\arraybackslash}X}
\newcolumntype{R}{>{\raggedleft\arraybackslash}X}
\newcolumntype{P}[1]{>{\raggedright\arraybackslash}p{#1}}

\usepackage{microtype}
\usepackage{url}
\usepackage{xcolor}
\definecolor{mycolor}{HTML}{FF6600}
\newcommand{\correct}[1]{\textcolor{mycolor}{#1}}
\newcommand{\imagesep}{\vspace*{-3pt}}

\usepackage{balance}
\usepackage{cite}
\title{Visually grounded learning of keyword prediction from untranscribed speech}
\name{Herman Kamper, Shane Settle, Gregory Shakhnarovich, Karen Livescu}
\address{Toyota Technological Institute at Chicago}
\email{\{kamperh, settle.shane, greg, klivescu\}@ttic.edu}

\usepackage{todonotes}

\begin{document}

\maketitle

\begin{abstract}
During language acquisition, infants have the benefit of visual cues to ground spoken language. Robots similarly have access to audio and visual sensors. Recent work has shown that images and spoken captions can be mapped into a meaningful common space, allowing images to be retrieved using speech and vice versa. In this setting of images paired with untranscribed spoken captions, we consider whether computer vision systems can be used to obtain textual labels for the speech. Concretely, we use an image-to-words multi-label visual classifier to tag images with soft textual labels, and then train a neural network to map from the speech to these soft targets. We show that the resulting speech system is able to predict which words occur in an \mbox{utterance---acting} as a spoken bag-of-words classifier---without seeing any parallel speech and text.  We find that the model often confuses semantically related words, e.g. ``man'' and ``person'', making it even more effective as a \textit{semantic} keyword spotter.
\end{abstract}

\section{Introduction}

Current automatic speech recognition (ASR) systems use supervised models trained on huge amounts of annotated resources.
In an effort to alleviate this dependence on labelled data, there is growing interest in methods that can learn from untranscribed speech~\cite{park+glass_taslp08,jansen+etal_icassp13,lee+etal_tacl15,versteegh+etal_sltu16,kamper+etal_taslp16}.
Here we consider the problem of grounding unlabelled speech when paired with images.
Annotating speech is expensive and sometimes impossible, e.g.\ for endangered or unwritten languages~\cite{besacier+etal_speechcom14}; grounding speech using co-occurring visual contexts could be a way to train systems in such low-resource scenarios~\cite{chrupala+etal_arxiv17}.  This setting is also relevant in robotics, where audio and visual signals can be combined for learning new commands~\cite{luo+etal_icvs08,sun+vanhamme_csl13,taniguchi+etal_advrob16}, and for understanding language acquisition in humans, who have access to visual cues for grounding~\cite{smith+yu_cognition08,thiessen_cogsci10,rasanen_speechcom12,frank+etal_acl14}.

Specifically, we are interested in the setting considered in~\cite{synnaeve+etal_nipsworkshop14,harwath+etal_nips16}, where natural images of scenes are paired with spoken descriptions, and neither the images nor speech are labelled.
Both~\cite{synnaeve+etal_nipsworkshop14} and~\cite{harwath+etal_nips16} used paired neural networks to map images and speech into a common semantic space where matched images and spoken captions are close to each other.
This approach allows images to be retrieved using speech and vice versa.
The same task was also considered in earlier work on tagging mobile phone images with spoken descriptions~\cite{hazen+etal_interspeech07,anguera+etal_icmir08}.
Despite the practical relevance, and interesting extensions in follow-on work~\cite{chrupala+etal_arxiv17,harwath+glass_arxiv17}, this joint mapping approach does not give an explicit grounding of speech in terms of textual labels.

Here we consider the possibility of using externally trained computer vision systems, which do have access to textual labels, to provide (noisy) supervision for untranscribed speech.
Concretely, we use an external image-to-words multi-label visual classifier, predicting for an image a set of words that refer to aspects of the scene.
Using soft labels (probabilities) from this vision system, we train a convolutional neural network to map spoken captions to these soft unordered word targets.
The result is a speech model that can predict which words (from a fixed vocabulary defined by the vision system) occur in a spoken utterance---acting as a spoken bag-of-words (BoW) classifier.

The previous work in this setting~\cite{synnaeve+etal_nipsworkshop14,harwath+etal_nips16,harwath+glass_arxiv17,chrupala+etal_arxiv17} also makes use of intermediate features from pretrained vision models.
Our approach can be seen as a further way to exploit vision systems, by also using their textual classification output.

We first apply our word prediction model to two tasks: BoW prediction, where the aim is to predict an unordered set of words that occur in a given utterance, and keyword spotting, where the task is to retrieve all utterances in a collection that contain a given textual keyword.
Promising results are achieved on both tasks.
Analysis shows that many of the model errors are semantically related to the correct labels, e.g.\ the model retrieves the speech utterance ``a dog runs in the grass'' for the textual keyword ``field''.
These ``errors'' may be desirable in certain settings. So in a final task, we evaluate our model as a \textit{semantic} keyword spotter, where it achieves performance much closer to that of an oracle model trained using ground-truth transcriptions.

\vspace*{-1pt}
\section{Related work}

Our work intersects with several other research directions.
Recent studies have shown that using extra visual features from the scene in which the speech occurs can improve conventional ASR~\cite{sun+etal_slt16,gupta+etal_icassp17}.
These
systems still rely on labelled speech data, while our aim is to use vision to ground \textit{untranscribed} speech.
There has also been much interest in developing speech models that, instead of exact transcriptions, can learn from very noisy labels~\cite{aimetti+etal_interspeech10,renkens+vanhamme_interspeech15,duong+etal_naacl16,bansal+etal_acl17,palaz+etal_interspeech16}.
The study of~\cite{palaz+etal_interspeech16} particularly influenced our approach, since they build a speech system using textual BoW labels ($\S$\ref{sec:model_overview}).
In the vision community, image captioning has received much recent attention, where the goal is to  produce a fluent and informative natural language description for a visual scene~\cite{vinyals+etal_cvpr15,karpathy+feifei_cvpr15,fang2015captions,donahue+etal_cvpr15}. 
In natural language processing, images have also been used to capture aspects of meaning (semantics) of written language; see~\cite{silberer_phd15,bernardi+etal_jair16} for reviews.
Other studies have considered multimodal modelling of sounds (not speech) with text and images~\cite{owens+etal_eccv16,aytar+etal_nips16,vijayakumar+etal_arxiv17}, and phonemes with images~\cite{gelderloos+chrupala_coling16}.
\begin{figure}[!t]
    \centering
    \includegraphics[width=0.95\linewidth]{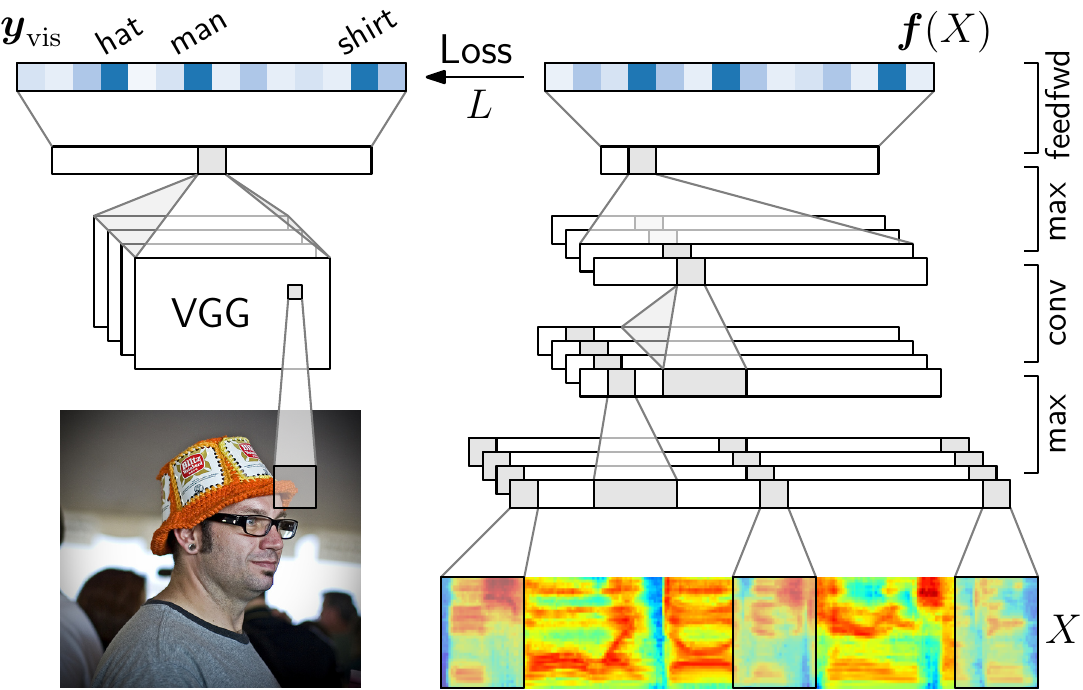}
    \imagesep
    \caption{A multi-label visual classifier is used to produce targets for training a word prediction model using only parallel images and unlabelled spoken captions.}
    \label{fig:speechvision}
    \vspace*{-8pt}
\end{figure}

\vspace*{-1pt}
\section{Word prediction from images and speech}
\label{sec:model}

Given a corpus of parallel images and spoken captions, neither with textual labels, we propose a method to train a spoken word prediction model using labels obtained from the visual modality.

\subsection{Model overview}
\label{sec:model_overview}

Every training image $I$ is paired with a spoken caption of frames $X = \vec{x}_1, \vec{x}_2, \ldots, \vec{x}_T$ (e.g.\ MFCCs).
We use a vision system to tag $I$ with soft textual labels, giving targets to train the speech network $\vec{f}(X)$ to predict which words 
are present in $X$.  The network $\vec{f}(X)$ therefore acts as a spoken {bag-of-words} (BoW) classifier (disregarding the order and quantity of words).
No transcriptions are used during training.
When applying the trained $\vec{f}(X)$, only speech input is used (and no image).
The approach is illustrated in Figure~\ref{fig:speechvision}, and below we give complete~details.

If we knew which words occur in training utterance $X$, we could construct a multi-hot vector $\vec{y}_{\textrm{bow}} \in \{0,1\}^W$, with $W$ the vocabulary size, and each dimension $y_{\textrm{bow}, w}$ a binary indicator for whether word $w$ occurs in $X$.
In~\cite{palaz+etal_interspeech16}, transcriptions were used to obtain this type of ideal BoW supervision.
Instead of a transcription for $X$, we only have access to the paired image $I$.
We use a multi-label visual classifier (with parameters $\vec{\gamma}$) which, instead of binary indicators, produces soft targets $\vec{y}_{\textrm{vis}}  \in [0,1]^W$, with $y_{\textrm{vis}, w}  = P(w | I, \vec{\gamma})$ the probability of word $w$ being present given image $I$.
In Figure~\ref{fig:speechvision}, $\vec{y}_{\textrm{vis}}$ would ideally be close to $1$ for $w$ corresponding to words such as ``hat'', ``man'' and ``shirt'', and close to $0$ for irrelevant dimensions.
This vision system is fixed:
during training (below), vision parameters $\vec{\gamma}$ are never updated.

Given $\vec{y}_{\textrm{vis}}$ as target, we train the word prediction model $\vec{f}(X)$.
This model (with parameters $\vec{\theta}$) consists of a convolutional neural network (CNN) over the speech $X$, as shown on the right in Figure~\ref{fig:speechvision}.
We interpret each dimension of the output as $f_w(X) = P(w | X, \vec{\theta})$.
Note that $\vec{f}(X)$ is not a distribution over the vocabulary, since any number of terms in the vocabulary can be present in an utterance; rather, each dimension $f_w(X)$ can have any value in $[0,1]$.  We train this speech network using the cross-entropy loss, which (for a single training example) is:
\vspace*{-5pt}
\begin{align}
    L(\vec{f}(X), \vec{y_\textrm{vis}}) 
    &= -\sum_{w = 1}^W \left\{ y_{\textrm{vis}, w} \log f_w(X) \;\; +\right. \nonumber \\ 
    &\qquad\qquad \left. (1 - y_{\textrm{vis}, w}) \log\left[1 - f_w(X) \right] \right\}  
    \label{eq:sigmoid_cross_entropy}
\end{align}
If we had $y_{\textrm{vis}, w} \in \{0, 1\}$, as in $\vec{y}_\textrm{bow}$, this could be described as the summed log loss of $W$ binary classifiers.
The size-$W$ vocabulary of our system
is implicitly specified by the vision~system.

\subsection{Two convolutional architectures over speech}
\label{sec:speech_networks}

We consider two different convolutional architectures for $\vec{f}(X)$.
Both deal with the variable number of frames in $X$ by pooling over the entire output of their final convolution layer.
As input layer, both use a one-dimensional convolution only over time, covering a number of frames and the entire frequency axis.

The first architecture is shown schematically in Figure~\ref{fig:speechvision}.
It is a CNN based on~\cite{harwath+etal_nips16,kamper+etal_icassp16}, consisting of several convolution and max pooling layers (final pooling covering the entire output), followed by fully connected layers.
A sigmoid activation is used for the final output $\vec{f}(X)$, and ReLUs in intermediate~layers.

\begin{figure}[!t]
    \centering
    \includegraphics[width=0.95\linewidth]{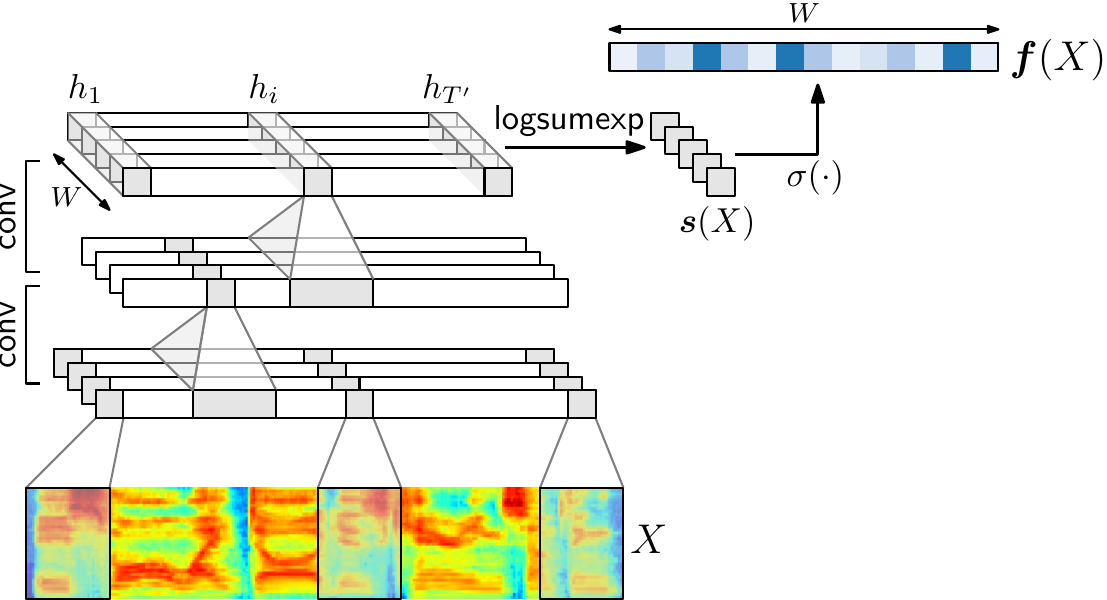}
    \imagesep
    \caption{A two-layer  \textbf{P}alaz, \textbf{S}ynnaeve and \textbf{C}ollobert (PSC) network ~\cite{palaz+etal_interspeech16}. The rest of our approach is as in Figure~\ref{fig:speechvision}.}
    \label{fig:psyc}
    \vspace*{-9pt}
\end{figure}

The second architecture is the one from \textbf{P}alaz, \textbf{S}ynnaeve and \textbf{C}ollobert~\cite{palaz+etal_interspeech16}, referred to as PSC.
It was originally developed for ideal BoW supervision ($\S$\ref{sec:model_overview}), with the aim of not only doing spoken BoW classification, but also locating where words occur in the speech.
PSC aims to do this by explicitly building the vocabulary into its final convolutional layer, as illustrated in Figure~\ref{fig:psyc}.
The final convolution is linear with $W$ output filters, matching the system vocabulary.  The outputs of these final filters are $\vec{h}_1, \vec{h}_2, \ldots, \vec{h}_{T'}$, with $\vec{h}_i \in \mathbb{R}^W$.
The idea is that $h_{i,w}$
gives a score for word $w$ occurring in the time span corresponding to output $i$, thus giving an estimate of where $w$ would occur in $X$.
To obtain the network output $\vec{s}(X) \in \mathbb{R}^W$, PSC does not use mean or max pooling, but rather an intermediate option:
\vspace*{-5pt}
\begin{equation}
    s_w(X) = \frac{1}{r} \log \left[ \frac{1}{T'} \sum_{t = 1}^{T'} \exp \left( r \, h_{t, w}(X) \right)  \right]
    \label{eq:psyc_pooling}
    \vspace*{-5pt}
\end{equation}
with $s_w(X)$ giving an overall unnormalized score for word $w$ being present in $X$.
This logsumexp pooling is equivalent to mean pooling when $r \rightarrow 0$ and max pooling for $r \rightarrow \infty$; Palaz {\it et al.}\ note that this intermediate method improves PSC's location prediction capability (refer to~\cite{palaz+etal_interspeech16}).
The final output of the network is $\vec{f}(X) = \sigma(\vec{s}(X))$, with $\sigma$ the sigmoid function.
We use $r = 1$, ReLU activations, and no intermediate pooling.

\subsection{The vision system}
\label{sec:vision_system}

In image captioning, the goal is to produce a natural language description of a scene~\cite{vinyals+etal_cvpr15,karpathy+feifei_cvpr15,donahue+etal_cvpr15,fang2015captions}. 
In contrast, rather than a fluent sentence, here we want a vision tagging system~\cite{barnard2003matching,guillaumin2009tagprop,chen2013fast} that predicts an unordered set of words (nouns, adjectives, verbs) that accurately describe aspects of the scene (Figure~\ref{fig:speechvision}, left).
This is a multi-label binary classification task, where for each word we must predict whether it is appropriate for the image.

We train our vision tagging system on the Flickr30k data set~\cite{young+etal_tacl14}, which contains $30\textrm{k}$ images, each with a set of $5$ captions, which we convert into a BoW after removing stop words. Given a limited set of task-specific training data, such as Flickr30k, a common approach is to start with a visual representation learned as a part of end-to-end training on a larger data set (possibly for a different task), and then adapt it to the task at hand.  We follow the established practice of using a representation trained for the ImageNet classification task~\cite{deng+etal_cvpr09}, as also in prior work~\cite{harwath+etal_nips16,chrupala+etal_arxiv17}.\footnote{The ImageNet output itself is not well-suited to our setting, since it performs a single multi-way classification among a set of image classes.}
Specifically, we use VGG-16~\cite{simonyan+zisserman_arxiv14}, but replace the final classification layer with four 3072-unit ReLU layers, followed by a binary classifier for word occurrence.
We train this multi-label visual classifier (with parameters $\vec{\gamma}$) on Flickr30k, with the output layer limited to the $W = 1000$ most common word types in the image captions.
The VGG-16 parameters are fixed during training; only the final layers that we add on top are trained.

Note that we train the vision system here only on Flickr30k images that do not correspond to train or test instances from the parallel image-speech data used in our experiments ($\S$\ref{sec:experiments}).
This leaves around $25\textrm{k}$ images.\footnote{We do this since there are, unfortunately, some overlapping images.} 
Also note that the vision system is trained and then fixed (parameters $\vec{\gamma}$ is not updated in $\S$\ref{sec:experiments}).

\vspace*{-1pt}
\section{Experiments}
\label{sec:experiments}

\subsection{Experimental setup}

We train our word prediction model on the data set of parallel images and spoken captions of~\cite{harwath+glass_asru15}, containing $8000$ images with $5$ spoken captions each.
The audio comprises around $37$~hours of active speech.
The data comes with train, development and test splits containing $30\,000$, $5000$ and $5000$ utterances, respectively. 
Speech is parametrized as MFCCs with first and second order derivatives, giving $39$-dimensional input.\footnote{We also tried filterbanks; MFCCs always worked similarly or better.}
Utterances longer than $8$~s are truncated ($99.5$\% of utterances are shorter than $8$~s).

Training images are passed through the vision system ($\S$\ref{sec:vision_system}), producing soft targets $\vec{y}_\textrm{vis}$ for training the word prediction model $\vec{f}(X)$ on the unlabelled speech.
We consider two architectures for $\vec{f}(X)$, referred to as VisionSpeechCNN and VisionSpeechPSC, respectively (see $\S$\ref{sec:speech_networks}).
VisionSpeechCNN is structured as follows: $1$-D ReLU convolution with $64$ filters over $9$ frames; max pooling over $3$ units; $1$-D ReLU convolution with $256$ filters over $10$ units; max pooling over $3$ units; $1$-D ReLU convolution with $1024$ filters over $11$ units; max pooling over all units; $4096$-unit fully-connected ReLU; and the $1000$-unit sigmoid output.
VisionSpeechPSC is structured as follows: $1$-D ReLU convolution with $96$ filters over $9$ frames; four $1$-D ReLU convolutions, each with $96$ filters over $10$ units; $1$-D linear convolution with $W = 1000$ filters over $10$ units; and logsumexp pooling followed by the final sigmoid activation.
We arrived at these two structures starting from those in~\cite{harwath+etal_nips16} and~\cite{palaz+etal_interspeech16}, respectively, and then tuned them on our development data.

We also obtain upper and lower bounds on performance.  As an upper bound, we train two oracle models, OracleSpeechCNN and OracleSpeechPSC, 
with the same structures as the two VisionSpeech models above.
These models are trained on ideal BoW supervision ($\S$\ref{sec:model_overview}): we obtain $\vec{y}_\textrm{bow}$ targets for the $1000$ most common words in the transcriptions of the $30\,000$ speech training utterances, after removing stop words.
Next, as a lower-bound baseline, we use a unigram language model prior that gives the unigram probability of each keyword as estimated from the transcriptions.
This baseline gives an indication of how much better our models do than simply hypothesizing common words.
Note that the textual transcriptions are used \textit{only} for the baseline and oracle models and for evaluation: neither of   the VisionSpeech models ever see any parallel speech and text.

All models were implemented in TensorFlow~\cite{tensorflow15}.\footnote{The code recipe is available at: {\scriptsize\url{https://github.com/kamperh/recipe_vision_speech_flickr}}.}
Based on development tuning, we use Adam optimization~\cite{kingma+ba_arxiv14} with a learning rate of $0.0001$ for all models, except those based on PSC, which uses $0.001$.

\subsection{Spoken bag-of-words prediction}

We first consider the task of predicting which words are present in a given test utterance.
Given input $X$, our model gives a score $f_w(X) \in [0, 1]$ for every word $w$ in its vocabulary, and these can be used for spoken BoW prediction.
To make a hard prediction, we set a threshold $\alpha$ and output labels for all $w$ where $f_w(X) > \alpha$.
We compare the predicted BoW labels to the true set of words in the transcriptions, and calculate precision, recall and $F$-score across \textit{all} word types in the reference transcriptions (not only the $1000$ words in the system vocabulary).  To compare performance independently of $\alpha$, we report average precision (AP), the area under the precision-recall curve as $\alpha$ is varied.

\begin{table}[!t]
    \mytable
    \caption{Spoken bag-of-word prediction performance (\%) at two thresholds $\alpha$, and the average precision (AP) over all $\alpha$.}
    \vspace*{-7.5pt}
    \begingroup
    \eightpt
    \begin{tabularx}{\linewidth}{@{}lRR@{\ \ \ }R@{\ \ \ }RR@{\ \ \ }R@{\ \ \ }R}
        \toprule
        & & \multicolumn{3}{c}{$\alpha = 0.4$} & \multicolumn{3}{c}{$\alpha = 0.7$} \\
        \cmidrule(l){3-5} \cmidrule(l){6-8} 
        Model & AP & $P$ & $R$ & $F$ & $P$ & $R$ & $F$ \\
		\midrule
        Unigram baseline & $6.8$ & $12.1$ & $14.2$ & $13.1$ & $17.6$ & $5.9$ & $8.8$ \\
        \addlinespace
        VisionSpeechCNN & $20.0$ & $34.4$ & $24.1$ & $28.3$ & $62.9$ & $8.9$ & $15.7$ \\
        VisionSpeechPSC & $18.9$ & $40.1$ & $20.2$ & $26.9$ & $62.9$ & $6.7$ & $12.0$ \\
        \addlinespace
        OracleSpeechCNN & $59.5$ & $78.3$ & $50.5$ & $61.4$ & $90.1$ & $43.5$ & $58.7$ \\
        OracleSpeechPSC & $69.7$ & $77.1$ & $63.0$ & $69.3$ & $87.4$ & $54.1$ & $66.8$ \\
        \bottomrule
    \end{tabularx}
    \endgroup
    \label{tbl:bow_prediction}
    \vspace*{-2pt}
\end{table}

Table~\ref{tbl:bow_prediction} presents BoW prediction performance for the different models at two operating points for $\alpha$, to show the trade-off between precision and recall.
The unigram baseline achieves non-trivial performance, indicating that 
some words are commonly used across the utterances in the data set.
Both VisionSpeech models substantially outperform this baseline at both $\alpha$'s, and in AP.
Although the VisionSpeech models still lag far behind the two oracle models, the VisionSpeech models are trained without seeing any parallel speech and text.
The precision of $61.3$\% of VisionSpeechCNN at $\alpha = 0.7$ is therefore noteworthy, since it shows that (although we miss many words in terms of recall), a relatively high-precision textual labelling system can be obtained using only images and unlabelled speech.
For the oracle models, the PSC architecture is beneficial, outperforming its CNN counterpart by all measures;
but for the VisionSpeech models, the PSC model falls slightly behind. We discuss this below.

\begin{table}[!t]
    \mytable
    \renewcommand{\arraystretch}{1.3}
    \caption{Example input utterances and BoW predictions of VisionSpeechCNN for $\alpha = 0.7$. \correct{Orange} shows correct predictions.}
    \vspace*{-7.5pt}
    \begingroup
    \eightpt
    \begin{tabularx}{\linewidth}{@{}LP{3cm}@{}}
        \toprule
        Transcription of input utterance & Predicted BoW labels \\
        \midrule
        a little girl is climbing a ladder & child, \correct{girl}, \correct{little}, young \\
        a rock climber standing in a crevasse & climbing, man, \correct{rock}\\
        man on bicycle is doing tricks in an old building & \correct{bicycle}, bike, \correct{man}, riding, wearing \\
        a dog running in the grass around sheep & \correct{dog}, field, \correct{grass}, \correct{running} \\
        a man in a miami basketball uniform looking to the right & \raggedright\arraybackslash ball, \correct{basketball}, \correct{man}, player, \correct{uniform}, wearing \\
        a snowboarder jumping in the air with a person riding a ski lift in the background &  \correct{air}, man, \correct{person}, snow, \correct{snowboarder}\\
        \bottomrule
    \end{tabularx}
    \endgroup
    \label{tbl:bow_prediction_examples}
    \vspace*{-9pt}
\end{table}

\begin{figure}[t]
    \centering
    \includegraphics[scale=0.5]{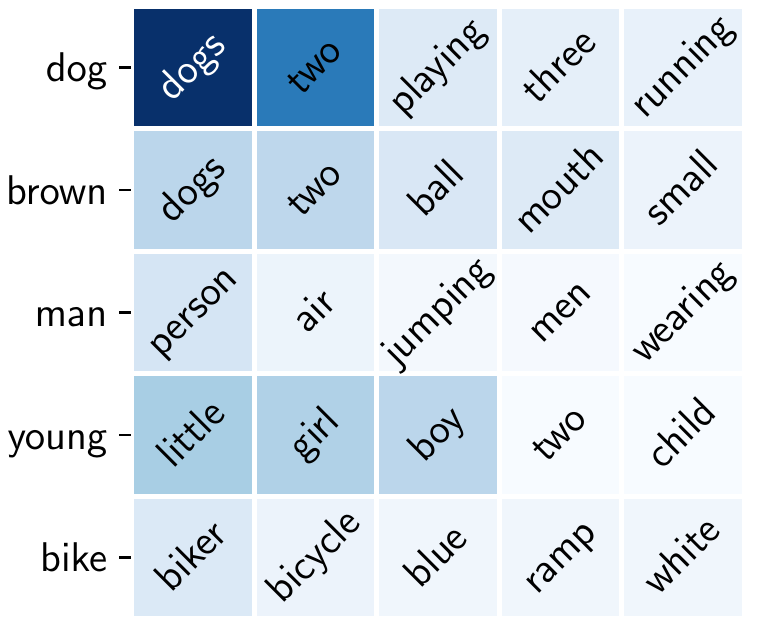} \hspace*{-5pt} \includegraphics[scale=0.5]{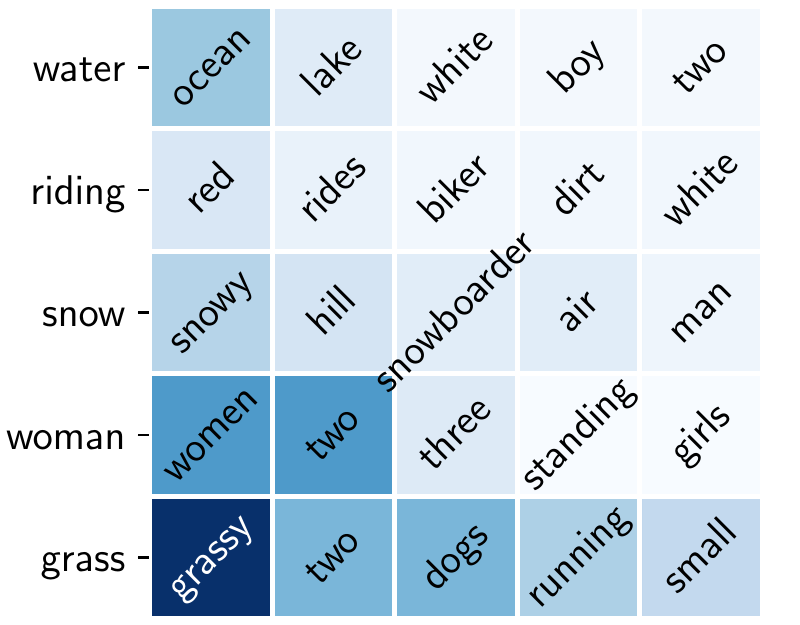}
    \caption{False alarms ($y$-axes) predicted by VisionSpeechCNN, and the words in the corresponding utterances (darker shading indicating higher frequency), when used for BoW prediction.}
    \label{fig:vision_speech_cnn_confusions}
\end{figure}

Table~\ref{tbl:bow_prediction_examples} gives examples of the type of output produced by the VisionSpeechCNN model.
To better analyze the model's behavior, we examine a selection of words that the model predicts that do not occur in the corresponding reference transcriptions.
Figure~\ref{fig:vision_speech_cnn_confusions} shows some of these ``false alarm words'', along with the most common words that do occur in the corresponding utterances.
In many cases, the predicted words are variants of the correct words: e.g.\ for an incorrect prediction of ``snow'', most of the reference transcriptions contain the word ``snowy''. Other confusions are semantic in nature, e.g.\ ``young'' is predicted when ``girl'' is present, and ``trick'' when ``ramp'' is present.

\subsection{Keyword spotting}

\begin{table}[t]
    \mytable
    \caption{Keyword spotting performance (\%).}
    \vspace*{-5pt}
    \begingroup
    \eightpt
    \begin{tabularx}{0.8\linewidth}{@{}lRRR}
        \toprule
        Model & $P@10$ & $P@N$ & EER \\
        \midrule
        Unigram baseline & $5.0$ & $3.5$ & $50.0$ \\
        \addlinespace
        VisionSpeechCNN & $54.5$ & $33.1$ & $22.3$ \\
        VisionSpeechPSC & $48.5$ & $31.9$ & $22.9$ \\
        \addlinespace
        OracleSpeechCNN & $92.0$ & $72.4$ & $6.2$ \\
        OracleSpeechPSC & $96.5$ & $83.0$ & $4.1$\\
        \bottomrule
    \end{tabularx}
    \endgroup
    \label{tbl:keyword_spotting}
    \vspace*{-4pt}
\end{table}

Our model can also be naturally used as a keyword spotter:
given a text query, the goal
is to retrieve all of the utterances in the test set containing spoken instances of that query.
We randomly select $20$ textual keywords from the VisionSpeech output vocabulary as queries.
For evaluation, we use three metrics~\cite{hazen+etal_asru09,zhang+glass_asru09}: $P@10$ is the average precision (across keywords, in \%) of the $10$ highest-scoring proposals; $P@N$ is the average precision of the top $N$ proposals, with $N$ the number of true occurrences of the keyword; and equal error rate (EER) is the average error rate at which the false acceptance and false rejection rates are equal.

Table~\ref{tbl:keyword_spotting} shows keyword spotting results.
The trend in relative performance is similar to that of Table~\ref{tbl:bow_prediction}: the unigram baseline performs worst, the VisionSpeech models give reasonable scores, and the oracle models perform best.
The very high oracle performance indicates that the constrained nature of the data used here (narrow domain, relatively small vocabulary) makes the task fairly easy when true transcriptions are available.
Nevertheless, it is again noteworthy that both VisionSpeech models obtain a $P@10$ of around $50$\% at an EER of $23$\%, without using any text.

To give a qualitative view of VisionSpeechCNN's errors, Table~\ref{tbl:keyword_error_examples} shows examples of incorrectly matched utterances for some keywords.
As before,
many of these erroneous
utterances contain either variants of the keyword (e.g.\ ``play'' and ``playing'') or are semantically related (e.g.\ ``young'' and ``little girl'').
Although these matches would seem reasonable and even desirable in some settings, they are penalized under the metrics in Table~\ref{tbl:keyword_spotting}.

\begin{table}[!t]
    \mytable
    \caption{Examples of incorrectly retrieved utterances when VisionSpeechCNN is used for keyword spotting.}
    \vspace*{-5pt}
    \begingroup
    \eightpt
    \begin{tabularx}{\linewidth}{@{}lLl@{}}
        \toprule
        Keyword & Example of incorrectly matched utterance & Type \\
        \midrule
        behind & a surfer does a flip on a wave & mistake \\
        bike & a dirt biker flies through the air & variant \\
        boys & two children play soccer in the park & semantic \\
        large & \dots a rocky cliff overlooking a body of water & semantic \\
        play & children playing in a ball pit & variant \\
        sitting & two people are seated at a table with drinks & semantic \\
        yellow & a tan dog jumping over a red and blue toy & mistake \\
        young & a little girl on a kid swing & semantic \\
        \bottomrule
    \end{tabularx}
    \endgroup
    \label{tbl:keyword_error_examples}
\end{table}

\subsection{Semantic keyword spotting}

To investigate this issue quantitatively, we considered the top $10$ proposed utterances for each keyword for each model, and relabelled as correct those utterances that either contained keyword variants or were semantically related.
This allows us to report $P@10$ for the task of \textit{semantic} keyword spotting, as shown in Table~\ref{tbl:semantic_keyword} (the other metrics would require us to semantically label all test utterances).
Compared to Table~\ref{tbl:keyword_spotting}, the semantic keyword spotting performance is better than exact keyword spotting scores for all models.
However, the VisionSpeech models improve most, with VisionSpeechCNN improving by almost 30\% absolute.
Moreover, while the oracle models improved mainly due to variant matches,
the VisionSpeech models had about equal numbers of relabelled variant and semantic matches.

In Tables~\ref{tbl:keyword_spotting} and~\ref{tbl:semantic_keyword} we again see that while PSC is superior to CNN for the oracle models, this is not the case for the VisionSpeech models.
As mentioned in $\S$\ref{sec:speech_networks}, PSC is intended to also estimate word locations.
Our results suggest that when trained on transcriptions (oracle models), there is a benefit in attempting to capture aspects of word order.
However, when trained through visual grounding, the output of VisionSpeechPSC produces high probabilities for several semantically related words, and there is far less structure in the order of these words.


\begin{table}[t]
    \mytable
    \caption{Semantic keyword spotting performance (\%)}
    \vspace*{-5pt}
    \begingroup
    \eightpt
    \begin{tabular}{lr}
        \toprule
        Model & $P@10$ \\
        \midrule
        Unigram baseline & $10.0$ \\
        \addlinespace
        VisionSpeechCNN & $82.5$ \\
        VisionSpeechPSC & $71.5$ \\
        \addlinespace
        OracleSpeechCNN & $98.0$ \\
        OracleSpeechPSC & $99.5$ \\
        \bottomrule
    \end{tabular}
    \endgroup
    \label{tbl:semantic_keyword}
\end{table}




\section{Conclusion}

We have introduced a new way of using images to learn from untranscribed speech.  By using a visual image-to-word classifier to provide soft labels for the speech, we are able to learn a neural speech-to-keyword prediction system. 
Our best model achieves a spoken bag-of-words precision of more than $60$\%, and a keyword spotting $P@10$ of more than $50$\% with an equal error rate of $23$\%.
The model achieves this performance without access to any parallel speech and text.
Further analysis shows that the model's mistakes are often semantic in nature, e.g.\ confusing ``boys'' and ``children''.
To quantify this, we evaluated our model as a \textit{semantic} keyword spotter, where the task is to find all utterances in a corpus that are semantically related to the textual keyword query.
In this setting, our model achieves a semantic $P@10$ of more than $80$\%.
Future work will consider how semantic search in speech can be formalized, and how the visual component of our approach can be explicitly tailored to obtain an improved visual grounding signal for unlabelled~speech.

\vspace{4pt}
{
\eightpt
\noindent\textbf{Acknowledgements:} We thank Gabriel Synnaeve and David Harwath for assistance with data and models, as well as Shubham Toshniwal and Hao Tang for helpful feedback.   This research was funded by NSF grant IIS-1433485. The opinions expressed in this work are those of the authors and do not necessarily reflect the views of the funding agency.
}

\newpage\balance
\bibliography{interspeech2017}

\end{document}